\documentclass{article}

\usepackage[dblblindworkshop, final]{neurips_2025}

\workshoptitle{AI for Science}

\usepackage[utf8]{inputenc} 
\usepackage[T1]{fontenc}    
\usepackage{hyperref}       
\usepackage{url}            
\usepackage{booktabs}       
\usepackage{amsfonts}       
\usepackage{nicefrac}       
\usepackage{microtype}      
\usepackage{xcolor}         
\usepackage{tabularx}
\usepackage{multirow}
\usepackage{tablefootnote}
\usepackage{enumitem}

\title{Reviewing Scientific Papers for Critical Problems\\With Reasoning LLMs:\\Baseline Approaches and Automatic Evaluation}

\author{%
  Tianmai M. Zhang\\
  University of Washington\\
  \texttt{tianmai@uw.edu} \\
  \And
  Neil F. Abernethy\\
  University of Washington\\
  \texttt{neila@uw.edu} \\
}

\begin{document}

\maketitle

\begin{abstract}
Recent advancements in large language models have sparked interest in utilizing them to aid the peer review process of scientific publication amid the peer review crisis. However, having AI models generate full reviews in the same way as human reviewers risks exacerbating the irresponsible use of LLM-generated reviews and instigating intentional manipulation. As an alternative, we propose adopting LLMs as manuscript quality checkers. We introduce several baseline approaches and an extendable automatic evaluation framework using top reasoning LLMs as judges to tackle the difficulty of recruiting domain experts for manual evaluation. Utilizing papers withdrawn from arXiv, we validated our proposed methods with several leading reasoning LLMs available in May-June 2025 and assessed their performance and API costs for identifying critical errors and unsoundness problems in scientific papers. o3 exhibited the best problem identification performance among all models at a modest cost. This paper provides insights into document-based scientific understanding/reasoning and lays a foundation for future applications. Our dataset, code, and model outputs are publicly available.
\end{abstract}

\section{Introduction}

Recent advancements in the domain intelligence of large language models (LLMs) have fostered interest in utilizing them to aid the peer review process of scientific publication, especially in consideration of the peer review crisis due to the skyrocketing number of paper submissions in recent years \citep{Kim25}. Researchers have reported receiving reviews that have likely been written with LLMs, and concern is growing over whether and how AI can be responsibly applied to aid peer review \citep{NatureNews25a, NatureNews25b}. The extensive knowledge and high efficiency of LLMs seem promising for streamlining laborious peer review, but irresponsible uses of LLM-generated reviews could significantly undermine trust in the long-established peer review process and publisher credibility.

Several studies have formally explored and evaluated the quality of LLM-generated reviews. An early study by \citet{Liang24} reported substantial overlap between GPT-generated reviews and human reviews, as well as participant-reported helpfulness of GPT-generated feedback. Later studies \citep{Du24, Zhou24, Shin25} revealed defects in LLM-generated reviews, such as superficial comments and the lack of criticism or novelty assessment. Other studies \citep{Darcy24, Gao24, Taechoyotin24, Tan24, Tyser24, Yu24, Zhu25, Zeng25} developed technical methods to improve LLM review generation. However, all of these studies focused on the scenario where LLMs generate full reviews in the same way as human reviewers, which risks exacerbating the irresponsible use of LLM-generated reviews and instigating intentional manipulation \citep{Ye24,Lin25}. The most common evaluation method in these studies was comparing LLM-generated reviews with human reviews, either manually or computationally.

Inspired by the Black Spatula Project\footnote{\url{https://the-black-spatula-project.github.io}} that seeks to identify errors in scientific papers, we propose using LLMs as manuscript quality checkers rather than efficient reviewers who write full reviews. In this way, LLMs would no longer be competing with human reviewers. Instead, LLMs would complete necessary sub-tasks, thereby saving reviewers' time.  This could allow human reviewers to focus more on leveraging their domain expertise to evaluate important aspects of the manuscript, such as completeness, coherence, novelty, and significance.

In this work, we consider the identification of critical errors and unsoundness problems that may invalidate the conclusions of a paper, a key sub-task in peer review, as the main goal of an LLM manuscript checker. We present an extensible framework including several baseline approaches and an automatic evaluation pipeline, which also supports other related tasks. Utilizing papers withdrawn from arXiv, we validate our proposed methods with several top-performing reasoning LLMs and assess their performance and costs to inform future research and applications.

\section{Methods}
\subsection{Dataset}

We utilized \textsc{WithdrarXiv} \citep{WithdrarXiv}, a large-scale dataset of papers withdrawn from arXiv by September 2024, along with associated retraction comments from authors and well-defined retraction categories. The most common retraction category, "factual/methodological/other critical errors in manuscript", contains 6,018 candidate cases with critical errors that would potentially invalidate study conclusions, such as flawed experimental designs, incorrect data analyses, and proof/lemma errors \citep{WithdrarXiv}.

Since retraction comments were typically short and did not necessarily contain clear mentions of errors (e.g., "\textit{This paper has been withdrawn due to some mistakes}"), we further filtered the dataset with the help of LLMs. Specifically, de-identified retraction comments were first provided to Gemini 2.5 Flash ($\texttt{preview-04-17}$) to determine whether each retraction comment clearly specified the error. This step resulted in a subset of 2,190 cases. Our manual review further excluded cases that (1) were incorrectly identified during LLM screening; (2) belonged to different versions of the same paper; (3) were not in English; (4) suspiciously used exactly the same retraction reason (e.g., "\textit{The author has withdrawn this paper due to a crucial sign error in equation 1}") that seems to have been a template provided by arXiv in early years; (5) contained problems that were unlikely to be detectable from the manuscript alone (e.g., an error in a core reference, a bug in code, unreliable raw data, problems revealed by new observations, etc.). We also corrected mistakenly redacted theorem names in the retraction comments. The final dataset, named \textsc{WithdrarXiv-Check}, contains 1,225 cases in total.

We randomly sampled 20\% of the dataset (245 cases) as the test set for evaluation experiments. The remaining 80\% (980 cases) of the dataset was set aside for training and validation, although these latter two steps were not considered in this work whose main objective was to establish baseline approaches and evaluation methods. Dataset characteristics are provided in Table~\ref{tab:dataset_characteristics}.

\begin{table}[ht]
  \centering
  \caption{Dataset characteristics}
  \begin{tabular}{lcc}
    \toprule
     & \textbf{Train} & \textbf{Test}\\
    \midrule
    \textbf{Sample size} & 980 & 245 \\
    \midrule
    \textbf{Time span}, n (\%)\\
    \quad 2007--2012 & 155 (16\%) & 32 (13\%)\\
    \quad 2013--2018 & 487 (50\%) & 114 (47\%)\\
    \quad 2019--2024 & 338 (34\%)& 99 (40\%)\\
    \midrule
    \textbf{Main subject}, n (\%)\\
    \quad Math & 492 (50\%) & 128 (52\%)\\
    \quad Physics\footnotemark & 256 (26\%) & 70 (29\%)\\
    \quad Computer Science & 196 (20\%) & 37 (15\%) \\
    \quad Others\footnotemark & 36 (4\%) & 10 (4\%)\\
    \midrule
    \textbf{Page count}\\
    \quad Median & 14 & 14\\
    \quad [Min, Max] & [1, 156] & [2, 136]\\
    \midrule
    \textbf{LaTeX script} & \multirow{2}{*}{--} & \multirow{2}{*}{216 (88\%)}\\
    \textbf{available}, n (\%)\\
    \bottomrule
  \end{tabular}
  \label{tab:dataset_characteristics}
\end{table}

\subsection{Baseline Approaches}

We propose three baselines for LLMs to perform quality checks based on different approaches to ingest the papers: (1) paper PDF as an attachment, (2) optical character recognition (OCR) results of the PDF in the prompt, and (3) LaTeX script of the paper in the prompt. The first approach simulates the common real-world scenario wherein users upload the PDF file of a paper as an attachment and ask an LLM to read it. However, the PDF preprocessing pipeline (and its performance) likely varies by LLM vendor.

The other two approaches based on OCR or LaTeX help decouple PDF preprocessing steps from LLM inference by providing the same input to all models, thereby enabling fairer comparisons of their competencies. However, LLM performance under the OCR approach could be limited considerably by OCR errors, especially when these errors appear in key math formulae, which are potentially more difficult to accurately transcribe. The LaTeX-based approach could perfectly retain the original math formulae, albeit with the inclusion of LaTeX markup, but may also introduce additional noise and information loss (e.g., page and section number). This method is also restricted to papers with available LaTeX source scripts.
\addtocounter{footnote}
{-1}\footnotetext{Contains physics, cond-mat, quant-ph, astro-ph, etc.}
\addtocounter{footnote}{+1}\footnotetext{Contains stat, q-bio, eess, econ, and q-fin.}

Considering the nature of the dataset, we evaluated the PDF-based approach and the LaTeX-based approach in this work. For the small proportion of papers without available LaTeX scripts (Table~\ref{tab:dataset_characteristics}), we resorted to utilizing the problems identified by the same model through the PDF-based approach. Images were ignored when using LaTeX scripts in the absence of a simple method to provide LLMs with images and correct image referencing at the same time\footnote{o-series models cannot read images in base64 format.}. For future work on papers in other scientific domains where PDF is the standard format for paper dissemination, we recommend testing the OCR-based approach as one of the baselines.

In our experiments, both approaches utilized the same simplistic, general task instruction (Appendix~\ref{sec:prompts}). In short, LLMs were instructed to produce a list of up to $k$ problems or errors that are the most critical in a given paper. The prompt was not customized for our dataset that is rich in math and physics papers. Each LLM checker was tested $n_c$ ($c$ for checker) times with each paper in consideration of potential variations in outputs.

\subsection{Evaluation}

Considering the daunting cost of recruiting domain experts to manually evaluate LLM-identified scientific errors, we propose an automatic evaluation pipeline to streamline the process. Inspired by LLM-as-a-judge and the idea of \textit{LLMs - You Can't Please Them All}\footnote{\url{https://www.kaggle.com/competitions/llms-you-cant-please-them-all}}, we utilized $m$ top-performing LLMs to serve as judges. Ideally, LLM judges should be from different LLM vendors to maximize preference diversity. Each LLM judge independently evaluates an LLM checker's problem submissions one by one for $n_j$ ($j$ for judge) times to determine whether they contain an exact match to the gold error description from the authors. For $n_j>1$, the most self-consistent answer of the judge is taken as its final decision. If an LLM checker receives a majority of (or all, for a stricter evaluation) affirmative votes from LLM judges, it is deemed to have made a hit on a paper. LLM checkers were primarily evaluated by their hit rates on test papers. Since hit rate is likely associated with $k$, the number of problems/errors allowed in generation, we report this metric as the Hit Rate at $k$ (HR@$k$). For $n_c>1$ (i.e., each LLM checker is tested on a paper more than once), the metric becomes the Mean Hit Rate at $k$ (MHR@$k$).

Another metric of interest from the application perspective is the proportion of true positives among all positive predictions, i.e., precision. In our task, all LLM-identified problems or errors are positive predictions, and an LLM checker with a higher precision would be more usable in real-world application. We again take the LLMs-as-judges approach, where each judge independently assesses $n_j$ times whether an individual problem submission is a true positive based on the paper in PDF, and a submission receiving a majority of (or all) affirmative votes is considered a true positive. In this way, we are able to obtain a rough estimate of the actual precision of LLM-identified problems from a paper. Please note that there is no gold standard for precision evaluation in our experiment, and that a case in which an LLM checker found no problem is skipped because a precision value cannot be calculated. LLM checkers were evaluated by their Average Precision (AP@ $k$) on test papers if $n_c=1$, or Mean Average Precision (MAP@ $k$) if $n_c>1$. Prompts for the judges can be found in Appendix~\ref{sec:prompts}.

\subsection{Experimental Setup}

In this work, we took $k=5$, $n_c=n_j=1$, and $m=2$, i.e., each LLM checker was tested once with each paper and was allowed to report up to 5 problems, and 2 LLMs served as judges, each judging a problem submission once.

The following reasoning LLMs were tested as paper quality checkers: Google's Gemini 2.5 Pro ($\texttt{preview-05-06}$) and Gemini 2.5 Flash ($\texttt{preview-04-17}$); OpenAI's o3 ($\texttt{2025-04-16}$) and o4-mini ($\texttt{2025-04-16}$); Anthropic's Claude 3.7 Sonnet ($\texttt{20250219}$).

The two LLM judges are Gemini 2.5 Pro ($\texttt{preview-06-05}$, for its better performance on related benchmarks than $\texttt{preview-05-06}$) and o3 ($\texttt{2025-04-16}$). We initially planned for $m=3$ with Claude 3.7 Sonnet as the last judge, but its overly low hit rate under the PDF-based approach indicates that it might not qualify for serving as a judge in this task. Under $m=2$, both judges must vote affirmatively to confirm a hit or a true positive.

LLMs were accessed via their official APIs using Python. Model parameters used are provided in Appendix~\ref{sec:model_parameters}. Reasoning effort or thinking budget was kept as the default or automatic setting if applicable. LLMs were not given access to any tools, including web search since LLMs might find the retraction comments online.

\section{Results}

Table~\ref{tab:results_all} provides the main results assessing LLM quality checkers.
Besides numbers of identified problems and performance metrics, we recorded token usage to inform future work that seeks to apply LLMs at a larger scale. Average costs of reviewing a paper under each pipeline-LLM combination were estimated based on the standard API pricing of the LLM vendors in June 2025.

Compared to the Gemini family, OpenAI's o-series models always tended to make full use of the 5 problem slots allowed. o3 achieved the highest estimated hit rate among all models at a modest cost. After switching to LaTeX, the hit rates of Gemini models decreased, likely due to the information loss. In contrast, the hit rates of OpenAI o-series models remained around the same or slightly increased, suggesting resistance to format change or a higher familiarity with LaTeX obtained in its training processes. When papers are provided as PDFs, Claude 3.7 Sonnet found no problem in 64.9\% of test papers, leading to a low hit rate compared to other reasoning models. Interestingly, both its number of identified problems and hit rate increased after switching to LaTeX, suggesting potential obstacles in Claude's PDF ingestion pipeline or scientific understanding via PDF.

Precision was assessed only under the PDF-based approach due to both its proximity to real-world usage and the availability of all information (especially figures and page/section numbers) in PDFs. Similar to the trend in hit rate results, larger models (Gemini 2.5 Pro and o3) received higher average precision estimates than their lightweight siblings. Gemini 2.5 Pro had a higher estimated precision than o3, likely due to its cautiousness in reporting problems. In terms of the total number of reported problems that were deemed true positives by both judges, the o3 checker found 350, Gemini Pro found 293, and Claude found only 149. This again suggests that o3 was the most competent problem finder in our evaluation, although additional verification steps are certainly necessary in future attempts to improve its precision.

\begin{table}[t]
  \centering
  \fontsize{9.4}{11}\selectfont
  \caption{Numbers of reported problems, performance, token usage, and costs of LLM checkers}
  \begin{tabular}{lcccccccc}
    \toprule
    \textbf{Checker Model} & \multicolumn{2}{c}{\textbf{\# Prob. Found}} & \multicolumn{2}{c}{\textbf{Performance} (\%)} & \multicolumn{3}{c}{\textbf{Avg. Token Usage} (/paper)} & \textbf{Cost Est.}\\
    \cmidrule(lr){2-3} \cmidrule(lr){4-5} \cmidrule(lr){6-8}
    & Avg. & Q1, Q3 & \textbf{HR@5} & \textbf{AP@5} & \textbf{Input} & \textbf{Think} & \textbf{Output} & (\$/paper)\\
    \midrule
    \multicolumn{9}{c}{\textit{PDF as an attachment}}\\
    \midrule
    Gemini 2.5 Pro & 3.3 & 3, 4 & 39.2 & 35.2 & 4,678 & 14,228 & 881 & 0.157\\
    Gemini 2.5 Flash & 3.8 & 3, 5 & 38.4 & 23.3 & 4,678 & 8,713 & 644 & 0.033\\
    o3 (medium) & 4.8 & 5, 5 & 48.2 & 29.5 & 16,594 & 3,152 & 729 & 0.064\\
    o4-mini (medium) & 4.7 & 5, 5 & 38.0 & 26.7 & 17,760 & 3,582 & 701 & 0.038\\
    Claude 3.7 Sonnet & 1.6 & 0, 4 & 11.0 & 36.4 & 43,357 & 1,630 & 311 & 0.159\\
    \midrule
    \multicolumn{9}{c}{\textit{LaTeX script in prompt}}\\
    \midrule
    Gemini 2.5 Pro & 3.3 & 3, 4 & 36.3 & -- & 20,644 & 18,069 & 1,033 & 0.217\\
    Gemini 2.5 Flash & 3.6 & 3, 5 & 34.7 & -- & 20,644 & 13,247 & 667 & 0.052\\
    o3 (medium) & 4.8 & 5, 5 & 50.6 & -- & 21,990 & 3,156 & 927 & 0.077\\
    o4-mini (medium) & 4.5 & 5, 5 & 38.8 & -- & 22,287 & 3,421 & 685 & 0.043\\
    Claude 3.7 Sonnet & 3.4 & 1, 5 & 24.1 & -- & 28,284 & 2,701 & 515 & 0.133\\
    \bottomrule
  \end{tabular}
  \label{tab:results_all}
\end{table}

Large differences in the input token usage between each model family reflect distinct PDF ingestion pipelines of the three LLM vendors. Unlike Gemini models that understand PDFs purely with vision\footnote{\url{https://ai.google.dev/gemini-api/docs/document-processing}}, OpenAI and Anthropic provide their LLMs with both the extracted text and an image of each page\footnote{\url{https://platform.openai.com/docs/guides/pdf-files?api-mode=responses}}\textsuperscript{,}\footnote{\url{https://docs.anthropic.com/en/docs/build-with-claude/pdf-support}}. It remains unclear why Claude has an exceptionally high input token usage for PDF files, and why o4-mini sometimes used slightly more input tokens than o3. After switching to unified LaTeX scripts, all models used the same magnitude of input tokens, although Claude still consumed more.
Regarding thinking token usage, Gemini models notably spent several times as many thinking tokens as o3 and o4-mini under both approaches, but this potentially overthinking behavior did not result in higher hit rates.

Additional evaluation results by individual judges are shown in Table~\ref{tab:results_by_judge}. Compared to performance scores determined by a single judge, final scores reasonably dropped due to the difficulty of receiving affirmative votes from both judges, which demonstrates some resistance of our multi-judge approach to potential false positives in automatic evaluation. The Gemini 2.5 Pro judge is consistently more lenient than o3, sometimes making hallucinatory assumptions about the relationship between a problem submission and the gold error description. For example, when comparing an LLM checker's report of a missing factor in an unnumbered equation after a specific sentence and a retraction comment regarding an error in equation 13, the Gemini 2.5 Pro judge said it is "extremely likely" that the erroneous equation is equation 13. In our tests with different judge instructions, Gemini 2.5 Pro was also far less responsive than o3 to important instructions such as "Default your answer to `No' and only give `Yes' if you are certain". The phrase "\textit{exactly} the same problem" in the prompt for judges to determine hits is particularly necessary for Gemini 2.5 Pro, without which it would give an even higher proportion of affirmative votes, although the trend in hit rate results remained the same. These observations reveal preference differences between models from different vendors and thus highlight the necessity of adopting multiple judges or model settings instead of a single judge while using the LLM-as-a-judge approach. 

\begin{table}[t]
  \caption{Evaluation results by individual judges}
  \centering
  \begin{tabular}{lcc}
    \toprule
    \multicolumn{3}{c}{\textbf{Hit Rate Evaluation}}\\
    \midrule
    \textbf{Checker Model} & \multicolumn{2}{c}{\textbf{Judge Model}}\\
    \cmidrule(lr){2-3}
    & Gemini 2.5 Pro & o3 (medium)\\
    \midrule
    \multicolumn{3}{c}{\textit{PDF as an attachment}}\\
    \midrule
    Gemini 2.5 Pro & 68.2 & 41.2\\
    Gemini 2.5 Flash & 66.5 & 40.4\\
    o3 (medium) & 80.4 & 50.2\\
    o4-mini (medium) & 72.7 & 40.4\\
    Claude 3.7 Sonnet & 24.1 & 12.2\\
    \midrule
    \multicolumn{3}{c}{\textit{LaTeX script in prompt}}\\
    \midrule
    Gemini 2.5 Pro & 64.5 & 38.0\\
    Gemini 2.5 Flash & 64.1 & 35.9\\
    o3 (medium) & 79.2 & 51.4\\
    o4-mini (medium) & 71.8 & 40.0\\
    Claude 3.7 Sonnet & 48.6 & 26.5\\
    \midrule
    \multicolumn{3}{c}{}\\
    \midrule
    \multicolumn{3}{c}{\textbf{Average Precision Evaluation}}\\
    \midrule
    \textbf{Checker Model} & \multicolumn{2}{c}{\textbf{Judge Model}}\\
    \cmidrule(lr){2-3}
    & Gemini 2.5 Pro & o3 (medium)\\
    \midrule
    \multicolumn{3}{c}{\textit{PDF as an attachment}}\\
    \midrule
    Gemini 2.5 Pro & 70.7 & 39.7\\
    Gemini 2.5 Flash & 46.4 & 29.8\\
    o3 (medium) & 55.0 & 36.9 \\
    o4-mini (medium) & 48.3 & 32.7\\
    Claude 3.7 Sonnet & 51.1 & 47.5\\
    \bottomrule
  \end{tabular}
  \label{tab:results_by_judge}
\end{table}

\section{Discussion}

This work introduces and validates a framework for automatic evaluation of LLMs for scientific quality checking. Evaluation results revealed impressive performance of leading reasoning LLMs in reviewing scientific papers for critical errors and unsoundness problems. We believe that our proposed design, after careful improvement and domain expansion, has the potential to serve as a new benchmark for document-based scientific understanding and reasoning. Our design is also generalizable to the detection of many other types of errors or problems that may appear in scientific papers, such as data errors, content inconsistencies, unmet publication requirements, and undeclared limitations, as long as gold standard annotations are available.

The idea of using natural language processing (NLP) for detecting flaws in scientific papers is, in fact, not completely new \citep{Kuznetsov24}. Previous studies have explored the application of NLP in the detection of statistical reporting inconsistencies \citep{Nuijten20}, mathematical or conceptual errors \citep{Liu23}, and inaccurate citations \citep{Sarol24, Zhang24}. There are also attempts in the industry to develop LLM-driven tools for similar purposes \citep{NatureNews25a}. The core novelty of our work is that we present a reusable framework for the formal evaluation of LLM capabilities in a wide range of specialized tasks related to scientific quality checking using real, full papers.

In this work, we only assessed the performance of the most simplistic approach---providing an LLM with a paper and tasking it with finding problems. More complex approaches or newer generations of models are likely to demonstrate further gains in LLM performance. Examples of these approaches include further customizing the prompt (e.g., by considering the scientific field a paper belongs to), designing a propose-then-verify workflow, expanding task input (e.g., including supplementary materials, paper references, or additional domain knowledge) and tools (e.g., code execution), fine-tuning models for specific reviewing tasks, or adopting multi-agent collaboration.

One potential concern over the validity of our evaluation results is that LLMs might have seen the test papers (or other versions of them) in their training data. However, recent studies on data contamination and membership inference attacks \citep{Duan24,Fu25} suggest that LLMs are unlikely to memorize individual instances from pre-training data due to the combination of few training epochs and enormous corpora. Reasoning LLMs also experienced complex fine-tuning and post-training processes, which could further obfuscate memorized information. Nevertheless, it would be beneficial to evaluate LLMs on papers that are unlikely to be included in the training phase, such as those published after the knowledge cutoff dates of LLMs.

Copyright restrictions pose a key challenge to the automatic review of scientific papers in multiple fields at scale. In this work, we circumvent this challenge by utilizing publicly available arXiv papers. In real-world applications of LLMs, researchers must consider confidentiality and legal risks before sending private, unpublished manuscripts or copyrighted papers to public interfaces provided by LLM service vendors. We also advise developers who wish to apply LLM review at scale to refrain from posting LLM-generated reviews online before performing careful examination for false positives.

Furthermore, we would like to restate our standpoint that human experts should always be at the center of peer review. Although our results demonstrate seemingly impressive capabilities of LLMs in finding critical problems in papers in the domains considered, it should not be interpreted as meaning that LLMs at the current stage are broadly competent to replace human reviewers. Instead, journal publishers and conference organizers may consider incorporating LLM quality checkers into initial assessments of manuscripts \citep{Bauchner24}, thereby reducing the burden on reviewers.

There can also be other ways of leveraging LLMs for peer review. For example, \citet{Kim25} proposed presenting LLM-generated reviews alongside human reviews to "potentially" discourage irresponsible human reviewers. However, these diverse approaches would likely necessitate separate lines of sociotechnical research with different task formulations, methodologies, and cost-effectiveness. We advocate that LLM-aided review processes should be implemented ethically and transparently \citep{Lin23, Zhuang25}, in cooperation with the wider scientific community to ensure that they result in a net increase in trust in the scientific literature. This is especially important at a time when AI tools are displacing human employees and the trust in science is at risk.

\section{Limitation}

As a proof-of-concept study, this work has several limitations. First, only closed-source reasoning LLMs were evaluated. Future work may consider comparing different PDF preprocessing pipelines and open-source LLMs. Second, our evaluation metrics may suffer inaccuracies due to (1) ambiguities in paper authors' retraction comments (e.g., "\textit{withdrawn due to a crucial error in Lemma 2}", without further specification of the error, (2) automatic evaluation utilizing LLMs without human involvement, and (3) our settings that each LLM checker was tested only once per paper and each LLM judge graded each submission only once. Nevertheless, a higher score under parallel evaluation still reasonably indicates better performance of an LLM checker. We expect gradual improvements in the reliability of the LLMs-as-judges approach as leading models continue to evolve. Third, the impacts of experiment parameters or alternate prompts were not formally investigated or optimized. In addition, our results based on a dataset rich in math and physics papers published in the past may not generalize well to papers in other scientific domains or future papers. Last, since we were unable to recruit domain experts for manual annotation, it remains possible that some cases in the dataset contain problems that could not be detected based on the manuscript alone, or that the authors gave inaccurate reasons for retraction, which may pose an upper limit for measurement of model performance using our dataset. If conditions permit, future work should also consider involving domain experts in calibration of LLM judges to resolve potential biases caused by circular evaluation of LLMs using LLMs.

A concurrent work on arXiv by \citet{Son25} also utilizes the \textsc{WithdrarXiv} dataset for automatic error detection in scientific papers. Unlike their work which focuses on benchmarking LLMs, our work approaches the same topic more from an application perspective. Their main metrics overlap substantially with our hit rate, although there is a difference in the meaning of $k$. Other key differences include:
\begin{itemize}[leftmargin=*]
\item Son et al. present a small but better annotated dataset of latest papers designed solely for one-time benchmarking, whereas our dataset is much larger and contains more papers from the past, also including a training set which may directly benefit future work.
\item They normalized all papers using OCR, whereas we assessed LLMs with papers in PDF and LaTeX formats.
\item They used GPT-4.1 to determine whether an LLM's error submissions match the gold error annotation, whereas we adopted multiple reasoning LLMs as judges.
\item They treated all LLM-identified problems that did not strictly match the gold error description as wrong answers, whereas we allowed some flexibility through LLM reasoning and attempted to assess the precision of these additional answers.
\end{itemize}
These factors could explain the gap between their and our evaluation results. The strengths of their study include the use of domain experts for further quality control of their benchmark dataset, inclusion of open-source LLMs in evaluation,  and a study of variations in LLM outputs via resampling. The two efforts are complementary, and both report the leading performance of o3.

\section*{Data Availability}
\label{data_availability}

Our \textsc{WithdrarXiv-Check} dataset, experiment code, and model outputs (including thinking outputs if available) are available on Github\footnote{\url{https://github.com/tianmai-zhang/WithdrarXiv-Check}}. Readers may use the dataset for further experiments, reproduce the numbers in our tables, or perform a closer inspection of the model outputs to obtain further insights into LLM behaviors.

\begin{ack}
This work is supported by the Ira Kalet and Fred Wolf Endowment Fund from the Department of Biomedical Informatics and Medical Education, University of Washington. The authors declare no competing interest.

The authors thank community members of the Black Spatula Project for their inspiring discussions, especially Ugo Dos Reis for highlighting the \textsc{WithdrarXiv} dataset in December 2024, Delip Rao for instructions on how to use this dataset, and Casey Wimsatt and David Reinstein for constructive feedback on early versions of this paper. The authors also thank Yuxin Xu for her help with the API.

\end{ack}

\bibliographystyle{plainnat}
\bibliography{custom}

\appendix

\section{Prompt Templates}
\label{sec:prompts}

\textbf{Prompt for LLM quality checkers:}\\
Please check the attached paper for critical errors and unsoundness problems that would invalidate the conclusions. You can ignore minor issues (e.g, typos and formatting errors) and limitations that have been properly acknowledged.\\
\\
In your final output, give me up to \{k\} most critical problems as a JSON object using the following schema: Entry = \{"Problem": str, "Location": str, "Explanation": str\}, Return: list[Entry]. For location, give page number, section number, equation number, or whatever applicable. You can end the list early if there are fewer problems. No need to provide references.\\
\\
\{paper\_attachment\_or\_LaTeX\_script\}\\
\\
\textbf{Prompt for LLM judges to determine hits}:\\
My colleague was reading a paper and said there is a problem in it, as described below:\\
Problem: \{problem\}\\
Location: \{location\}\\
Explanation: \{explanation\}\\
\\
I checked the paper and noticed that the authors have the following retraction comment:\\
\{retraction\_comment\}\\
\\
Is my colleague referring to exactly the same problem mentioned in the retraction comment? Your final answer should be "Yes" or "No". Default your answer to "No" and only give "Yes" if you are certain. You may explain your decision but please be concise.\\
\textbf{Prompt for LLM judges to determine true positives}:\\
My colleague was reading this paper and said there is a critical problem in it, as described below:\\
Problem: \{problem\}\\
Location: \{location\}\\
Explanation: \{explanation\}\\
\\
Is this problem a true problem or a false alarm? Please be careful because I don't want to get the authors into trouble by mistake. In your final answer, clearly indicate "Yes, it is a true problem" or "No, it is a false alarm". Make your best decision if you are unsure. You may explain your decision but please be concise.\\
\\
\{paper\_attachment\}

\section{Model Parameters}
\label{sec:model_parameters}

\textbf{Gemini 2.5 Pro and Gemini 2.5 Flash}:\\
thinking budget: default (automatic)\\
include thoughts: True\\
tools: [] \quad temperature: 0 \quad seed: 42\\
\\
\textbf{o3 and o4-mini}:\\
reasoning effort: defaults to "medium"\\
reasoning summary: "auto"\\
tools: []\\
temperature and seed: not supported\\
\\
\textbf{Claude 3.7 Sonnet}:\\
max tokens: 16,000\\
thinking type: "enabled"\\
thinking budget: 14,000\\
tools: []\\
temperature: 1 (required for thinking)\\
seed: not supported

\end{document}